# Enhanced Prediction of CAR T-Cell Cytotoxicity with Quantum-Kernel Methods

Filippo Utro[1], Meltem Tolunay[2], Kahn Rhrissorrakrai[1], Tanvi P. Gujarati[2], Jie Shi[3,4], Sara Capponi[3,4], Mirko Amico[2], Nate Earnest-Noble[2], Laxmi Parida[1]

[1] IBM Research, Yorktown Heights, NY, 10598 USA

[2] IBM Quantum, Almaden Research Center, San Jose, CA, 95120 USA

[3] IBM Almaden Research Center, San Jose, CA, 95120 USA

[4] Center for Cellular Construction, San Francisco, CA, 94158 USA

## Abstract

Chimeric antigen receptor (CAR) T-cells are T-cells engineered to recognize and kill specific tumor cells. Through their extracellular domains, CAR T-cells bind tumor cell antigens which triggers CAR T activation and proliferation. These processes are regulated by co-stimulatory domains present in the intracellular region of the CAR T-cell. Through integrating novel signaling components into the co-stimulatory domains, it is possible to modify CAR T-cell phenotype. Identifying and experimentally testing new CAR constructs based on libraries of co-stimulatory domains is nontrivial given the vast combinatorial space defined by such libraries. This leads to a highly data constrained, poorly explored combinatorial problem, where the experiments undersample all possible combinations. We propose a quantum approach using a Projected Quantum Kernel (PQK) to address this challenge. PQK operates by embedding classical data into a high dimensional Hilbert space and employs a kernel method to measure sample similarity. Using 61 qubits on a gate-based quantum computer, we demonstrate the largest PQK application to date and an enhancement in the classification performance over purely classical machine learning methods for CAR T cytotoxicity prediction. Importantly, we show improved learning for specific signaling domains and domain positions, particularly where there was lower information highlighting the potential for quantum computing in data-constrained problems.

# Introduction

Cellular therapy is the manifestation of reprograming cells for therapeutic purposes. These repurposed cells can sense a different variety of external stimuli, process the information, and initiate biological processes that enable a therapeutic response. This strategy has proven effective in treating different malignancies, demonstrating great potential to cure several diseases. For this reason, cell therapies are emerging as a new paradigm in medicine [1–3]. In the context of immune cell therapies, chimeric antigen receptors (CARs) are genetically engineered T-cell receptors designed to target specific antigens and activate a T-cell response[4,5]. The CAR extracellular domain identifies specific, tumor-associated antigens and the intracellular, co-stimulatory domains trigger the T-cell activation and immune response (Fig. 1A). The motifs forming the co-stimulatory domains recruit different signaling proteins that initiate and propagate the signal intracellularly. Therefore, these motifs, or the combination of them, can determine CAR T-cell phenotypes and performances, and enhance specific functions for clinical outcomes, such as tumor remission[6,7]. To optimize the design of CAR T-cells, several studies have focused on mutations in co-stimulatory domains resulting in increased functions[7–11]. However, these studies are limited to selected domains and receptors with known functions in T-cells.

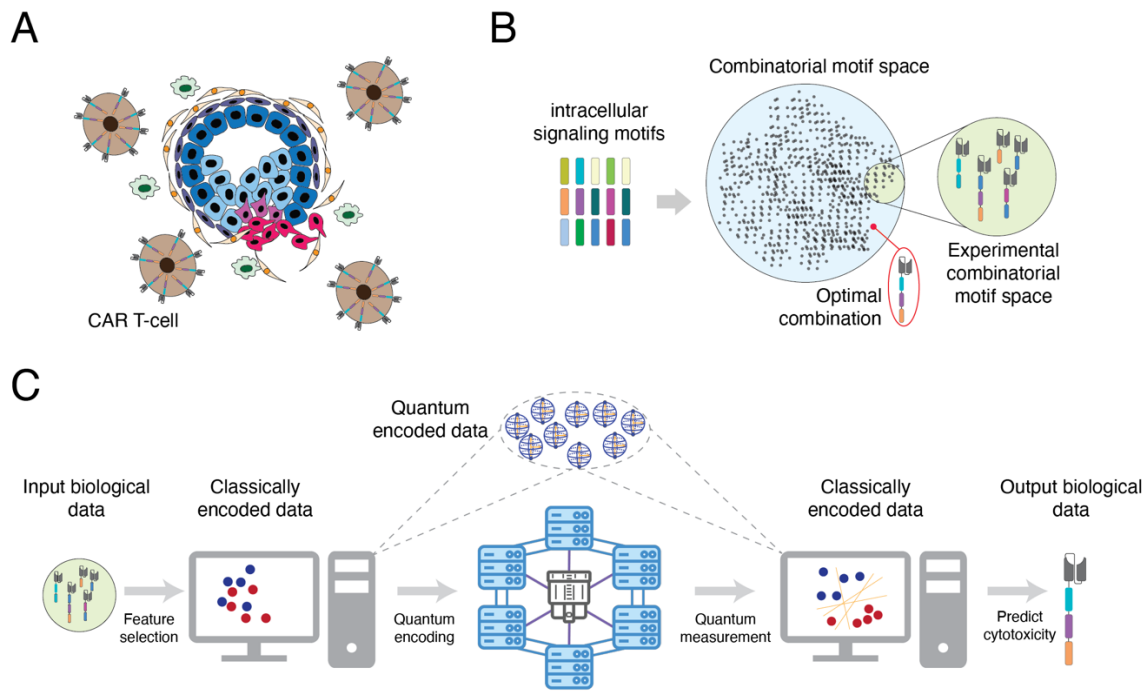

*Figure 1 **Application of the** projected quantum kernels (PQKs) algorithm to predict CAR T-cell cytotoxicity.* A) *Scheme of the Chimeric Antigen Receptors (CARs) T cell (brown cells) surrounding a mix of healthy (blue cells) and carcinogenic cells (red cells).* B) *The thirteen intracellular motifs can be combined in three different positions in the intracellular domain of the CAR, resulting in a large number of combinations. Each combination corresponds to a CAR T-cell with a specific phenotype. Experiments partially sample this large combinatorial space due to the high cost and production time of CAR T constructs and the optimal combination of co-stimulatory motifs related to the CAR T-cell with desired function might remain unsampled.* C) *Our PQK workflow operates by taking as input the experimentally generated combinatorial motif space and performing quantum encoding by loading it onto a quantum circuit, i.e. quantum embedding. This quantum-encoded data now lies in a high-dimensional Hilbert space, represented by qubits, and is projected back to a classical space through quantum measurements. This projected data is then analyzed with classical kernel methods to perform downstream tasks such as cytotoxicity prediction.*

Recently, a new approach to identify novel CAR constructs with desired functions has been used[12–15], which defines combinatorial libraries of co-stimulatory domains that are screened experimentally using state-of-the-art cell sorting and sequencing technologies. This approach yields CAR T-cell phenotypes that extend beyond those that can be generated by using native receptor domains. However, this approach is not optimized to extract nonlinear information from library screen data, which might contain details on interactions between co-stimulatory domains, relationships between these elements, and cell phenotypes. In addition, predictive models built using screen data might help to explore efficiently the combinatorial space defined by the co-stimulatory domains, which is often not feasible experimentally[16]. The limitations in time and effort to build and test new CAR constructs may result in a biased outcome because the optimal, desired construction might reside outside of the combinatorial sub-space tested experimentally (Fig. 1B).

Recently, an approach based on using a pooled library screening combined with classical machine learning (ML) methods including unsupervised clustering or principal components analysis was proposed[13–15]. Pooled screen methods allow the analysis of large combinatorial libraries but are less accurate in relating each phenotype with the specific combination of co-stimulatory domains because CAR T-cells may influence other CAR functions. Interestingly, to avoid this type of limitation and facilitate the identification of the optimal combination of signaling motifs, Daniels et al[12] used an arrayed screen approach combined with a predictive ML model based on deep learning architectures to guide the design of receptors with desired functions. Specifically, they built a combinatorial library based on the combinations of 13 motifs placed in 3 different positions (Fig. 1B) in the CAR intracellular domain. They tested experimentally only ~10% of the CAR T library, i.e. 246 experimental datapoints, and provided predictions on the entire combinatorial library by implementing a model based on a convolutional neural network (CNN) with a long-short term memory (LSTM) model trained on the small, limited experimental dataset. The CNN+LSTM model yielded an accuracy of $R^2=0.71$ on regressing the cytotoxicity and stemness data. However, given the clinical importance of a prediction to be highly accurate in the field of immunotherapy, models that can provide improved accuracy, particularly in settings with limited training data, could provide a significant step forward.

In the last decade, there has been significant progress in the development of quantum computing[17–19]. Quantum computing leverages features of quantum mechanics[20,21], such as quantum entanglement and quantum superposition, that could enable it to address problems that are extremely challenging for classical computing. There are numerous ongoing efforts to identify which problems may benefit with quantum computing approaches, for example within the areas of optimization[22,23], high energy physics[24], protein structure prediction[25], and healthcare and life sciences[26,27]. Given the initial success of classical ML for the problem of predicting CAR T-cell cytotoxicity, it is natural to explore the application of quantum machine learning (QML) [28–30] for this task. While there has been empirical evidence that certain QML algorithms, including quantum convolutional neural networks can generalize better on few data than classical ML[31], there are challenges for these variational algorithms to achieve exponential speedup on pre-fault tolerant quantum devices (PFTQDs), which are current quantum devices where instrumental noise is an important factor and whose errors are not completely corrected. These challenges also include the relatively limited number of available qubits and their connectivity, circuit execution times, and overhead associated with quantum state preparation[32–34] that may all require moving to a lower-dimensional feature or sample space. Yet there have been recent developments in hybrid quantum-classical ML approaches[35–37], which seek to bypass some of these challenges, enabling improvements in performance or discovery of novel insights that would not be possible with classical ML alone.

In this work we explore the use of one of these hybrid approaches, Projected Quantum Kernel (PQK)[37], to classify different CAR T-cell cytotoxicity phenotypes (Fig. 1C). We perform the largest scale demonstration to date for data classification with a digital quantum computer using PQK. Briefly, PQK uses a quantum feature map to embed classical data into a higher-dimensional feature space, i.e. a Hilbert space, using a quantum circuit. By taking measurements of local properties of the quantum state prepared by this circuit, this quantum embedding is returned to a different classical embedding which can then be analyzed using classical kernel methods, such as support vector machines (SVMs). Using PQK, we perform experiments with a 61-qubit quantum circuit to classify CAR T-cells for high and low cytotoxicity, demonstrating overall comparable performance to classical kernel methods while learning unique motif-specific signals from the data.

## Results

In this study we used a hybrid quantum-classical machine learning approach, PQK, to predict the cytotoxicity of CAR T-cell designs. We embed the high-dimensional input data into a quantum feature space and then project back into a lower-dimensional classical space (Fig 1C). After this projection step, we apply a classical kernel method, SVM, to this data tested with linear, radial basis function, sigmoid, and polynomial kernels. This hybrid approach leverages the power of quantum feature encoding while maintaining the computational efficiency of classical algorithms. We compare the classification accuracy of a classical SVM model with and without the PQK transformation of the data. In this way we can evaluate whether there is an improvement in predictive performance, or novel biological insights, that the PQK transformation is able to uniquely capture.

In the publication that introduced PQK[36], the authors propose a framework to assess the potential for quantum prediction advantage in a machine learning task. Accordingly, we first define a quantity called $g_{cq}$, which is a measure of the geometric separation between kernels constructed with classical and quantum-projected datasets.

$$g_{cq} = g(K^c \| K^q) = \sqrt{\left\| \sqrt{K^q}\sqrt{K^c}(K^c + \lambda I)^{-2}\sqrt{K^c}\sqrt{K^q} \right\|_\infty}$$

In the above equation, λ is the regularization parameter in the primal formalism of the SVM, and *K* are the classical and quantum-projected kernel matrices. If this quantity is on the order of √*N,* where *N* is the number of the training samples in the dataset (*N*=172), then we can move on to the second test to check for potential quantum prediction advantage. If $g_{cq}$ is significantly smaller than √*N*, then we can expect the classical model to perform as well as the quantum-projected model.

For the second test, we define additional metrics referred to as model complexities $s_c$ and $s_q$. A potential quantum prediction advantage can be expected in case $s_c$ is on the order of √*N,* while $s_q$ is smaller. Model complexity is defined as follows.

$$s_{K,\lambda}(N) = \sqrt{\frac{\lambda^2 \sum_{i=1}^{N}\sum_{j=1}^{N}(K+\lambda I)^{-2}_{ij} y_i y_j}{N}} + \sqrt{\frac{\sum_{i=1}^{N}\sum_{j=1}^{N}((K+\lambda I)^{-1}K(K+\lambda I)^{-1})_{ij} y_i y_j}{N}}$$

In the above equation, all symbols are as defined previously, and *y* are the true labels of the dataset. We performed an initial evaluation of our dataset with these metrics using the first two motif positions with binary encoding and eight repetitions of the ZZ Feature Map[38]. The results reveal a geometric separation between the classical and the quantum-projected radial basis function kernels of 15.777, which is similar to $\sqrt{N}$ = 13.115. In the second check, we observed the classical model complexity to be 6.090 vs 1.527 for the quantum-projected kernel, indicating the potential for PQK to outperform the classical model.

## Performance on Quantum Hardware

We provide empirical evidence of the benefits of using the PQK algorithm to transform the data and classify CAR T-cell designs for high or low cytotoxicity based on their signaling motifs using a quantum device. We process the data as discussed in the Method section and make use of all available features. For quantum embeddings, we explore two options: a ZZ feature map (E1) and a Heisenberg evolution circuit (E2)[37]. The same data is provided to the quantum and classical models. For the SVM, the 246 data points are directly input as a vector of length 60. For the PQK, a quantum circuit, i.e. feature map, is used to embed the input vector into 60 and 61 qubits for the E1 and E2 circuits, respectively. Following measurement of the circuit, the quantum projected data is then used by a SVM classifier whose hyperparameters are optimized using a grid search. Per data split, we report the maximum performance achieved by an optimized SVM on the projected data (Fig. 2 and Supp. Fig. 1). These experiments were carried out on a superconducting, gate-based quantum processing unit (QPU), an IBM Heron R2 QPU (*ibm_marrakesh*).

We compared the performance of PQK (E1) data against the original data using four different feature map repetition settings for the quantum embedding with $\frac{\pi}{2}$ or π rotation angle (Fig. 2A and Supp. Fig 1A). PQK-based classifiers were able to achieve higher median F1 scores using $\frac{\pi}{2}$ rotation angle compared to π and therefore we focused further analysis using this rotation angle. We observed that predictive performance is consistent across splits within a given repetition setting. As the number of feature map repetitions increases from 4 to 8, the median F1 score of the SVM classifier improves slightly using PQK data versus the original data, 0.75 vs 0.73, respectively. Moreover, at 8 feature map repetitions, the maximal F1 achieved over the 10 data splits was 0.81 for PQK data compared to 0.77 for the original data. The increase in performance with increasing circuit depth signals that the improvement in correlations between features afforded by increased interactions between qubits is beneficial to improve classification accuracy. Though when the repetition number increases to 12, PQK performance decreases, likely a consequence of the corresponding increase in the circuit depth. In fact, due to finite qubit lifetimes and gate fidelities, as the circuit depth increases the accumulation of noise on the quantum device will remove any further benefits from the higher dimensional feature space. The two-qubit gate depth increases from 16 layers of two-qubit gates at 4 feature map repetitions, to 48 layers of two-qubit gates at 12 repetitions (Table 1). When testing PQK with the Heisenberg embedding (E2), we achieve an equivalent median F1 to the original data, however increasing from four to six trotter evolution steps, the predictive performance worsened (Supp. Fig. 1B). Overall, these results demonstrate that with PQK (E1) data, an SVM is able to achieve marginally improved predictive performance on a PFTQD.

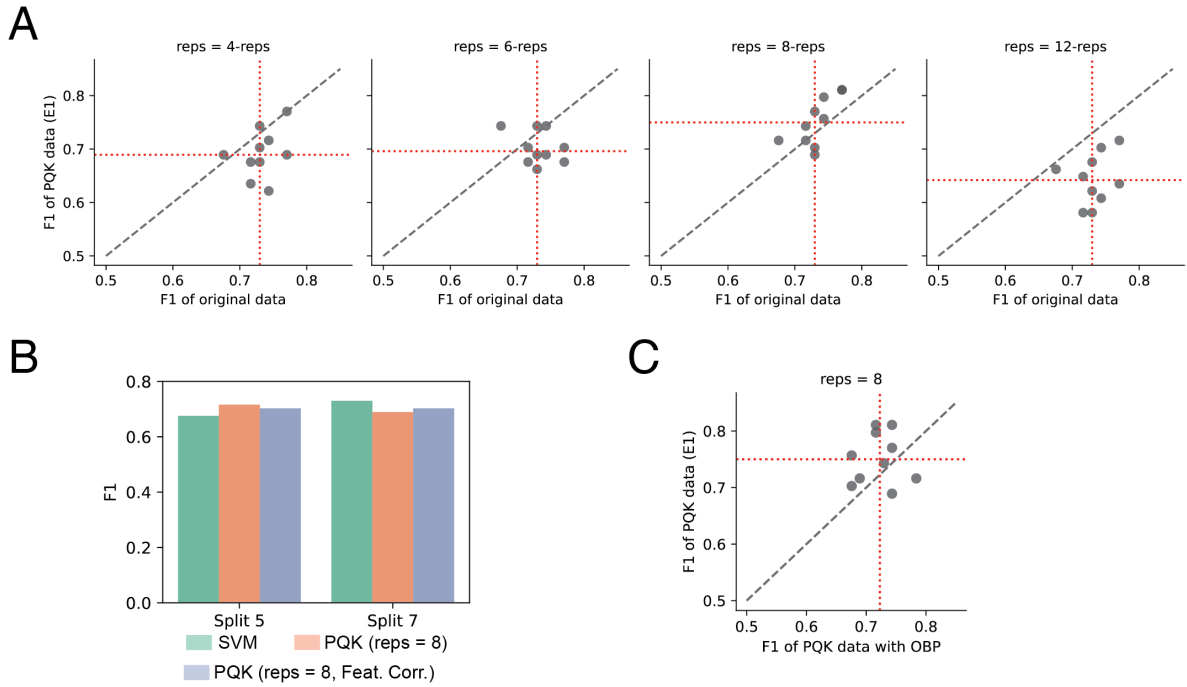

*Figure 2.* **Comparison of the predictive performance of PQK and SVM**. A) Predictive performance for PQK data (E1- with ZZ feature map) using angle rotation = $\frac{\pi}{2}$ is shown against an SVM with original data over 10 train/test splits. Each point corresponds to the F1 of the best model (as described in the Method Section) for a given split with PQK data on the y-axis and original on the x-axis. Red dotted lines indicate respective median values. B) Two data splits are shown with the prediction F1 shown with original data (green), PQK data (orange), and PQK data with features ordered by correlation (blue). The PQK (E1) uses 8 feature map repetitions and angle rotation = $\frac{\pi}{2}$. C) Comparison of predictive performance of PQK performed on QPU (y-axis) and in simulation with OBP (x-axis) using 8 feature map repetitions and angle rotation = $\frac{\pi}{2}$.

We then investigate whether the ordering of the features in the quantum circuit has an impact on the PQK model. Using pairwise entanglement in the quantum feature map, it is possible that highly correlated features are encoded to qubits that are far apart, may have reduced entanglement that does not sufficiently reflect this association. Prior to the quantum projection, we reordered the input features according to their correlation (see Method for details) and observed that the predictive performance was not meaningfully affected (Fig. 2B). This means that the circuit depth used suffices to introduce the needed correlation between features.

The quantum circuits implementing the PQK transformation are beyond the scale that can be simulated exactly even on the most powerful HPC machines. To compare the performance of a classical simulation of this transformation, we resort to approximate methods. In particular, we use Pauli backpropagation, or operator backpropagation (OBP)[39,40], of the measured observables to estimate the expectation values measured in the PQK transformation for each of the data point. Pauli backpropagation consists in decomposing the measured observable $\mathcal{O}$ in a linear combination of Pauli terms $\mathcal{O} = \sum_i c_i P_i$. Each gate, $G$, in the circuit is then applied to the observable (starting from the end of the circuit) to evolve it in the Heisenberg picture $\mathcal{O}' = G^\dagger \mathcal{O} G$. We adopt a coefficient-based truncation of the terms in the backpropagated observable to reduce the exponential growth in the number of terms in the observable due to the presence of non-Clifford gates. We tested different thresholds for the minimum value of the coefficient of each term – 0.2, 0.1, and 0.05 – until a convergence was observed. Compared to experimental runtimes on QPU, OBP experiments with a 0.2 threshold were six times slower and at the 0.05 threshold were nine times slower (on 198 core Intel(R) Xeon(R) Platinum 8260 CPU @ 2.40GHz with 1536 GB of RAM). We observed that the classification accuracy of PQK executed on a quantum device outperforms its approximate OBP simulation (at the highest accuracy setting of 0.05), achieving a median F1 of 0.73 (with maximal F1 over 10 splits of 0.78) versus 0.75 (with maximal F1 over 10 splits of 0.81) on QPU (Fig. 2C).

Table 1. Pre-transpiled E1 (60-qubit) and E2 (61-qubit) circuit sizes.

| Number of repetitions (E1) or trotter steps (E2) | E1 or E2 | Total number of gates | Total number of 2-qubit gates | 2-qubit gate depth |
|---|---|---|---|---|
| 4 | E1 | 1188 | 472 | 16 |
| 6 | E1 | 1782 | 708 | 24 |
| 8 | E1 | 2376 | 944 | 32 |
| 12 | E1 | 3564 | 1416 | 48 |
| 4 | E2 | 4141 | 1440 | 48 |
| 6 | E2 | 6181 | 2160 | 72 |

## PQK model captures unique signals

While the overall predictive performance for the PQK data is only marginally improved compared to using the original data, there remains the potential for PQK to bring orthogonal insights that may complement a purely classical framework. We consider whether a bias could be observed towards the PQK data to better predict CAR T-cell cytotoxicity with certain signaling motifs at a given position as compared to the original data (Fig. 3 and Supp. Figure 2-5). We observe that the classifier on the PQK data was able to consistently predict specific position and motif types which were not captured from the original data. There were no instances where the SVM on the original data was able to better predict consistently a specific motif and position combination over the 10 splits (Fig. 3). We also find that the frequency for which PQK data are able to do significantly better than the original data is greatest in motif position three, which is the position with the fewest datapoints. This position is more likely to be empty or have a terminating motif than the other two positions and would therefore have the least amount of information on which to learn. For example, the PQK-based classifier was able to predict CAR T-cells with motifs from GAB1 significantly better than SVM with the original data at only the third position (Fig. 3F), yet was able to predict CD40, LAT, and IRAK1 derived motifs better than SVM with original data at all three positions (Fig. 3D-F). The number of source proteins that see significant benefit from using PQK data rather than the original data increases from 3 to 6 to 8 moving from position 1 to 3, with a similar trend for binding partners and motif.

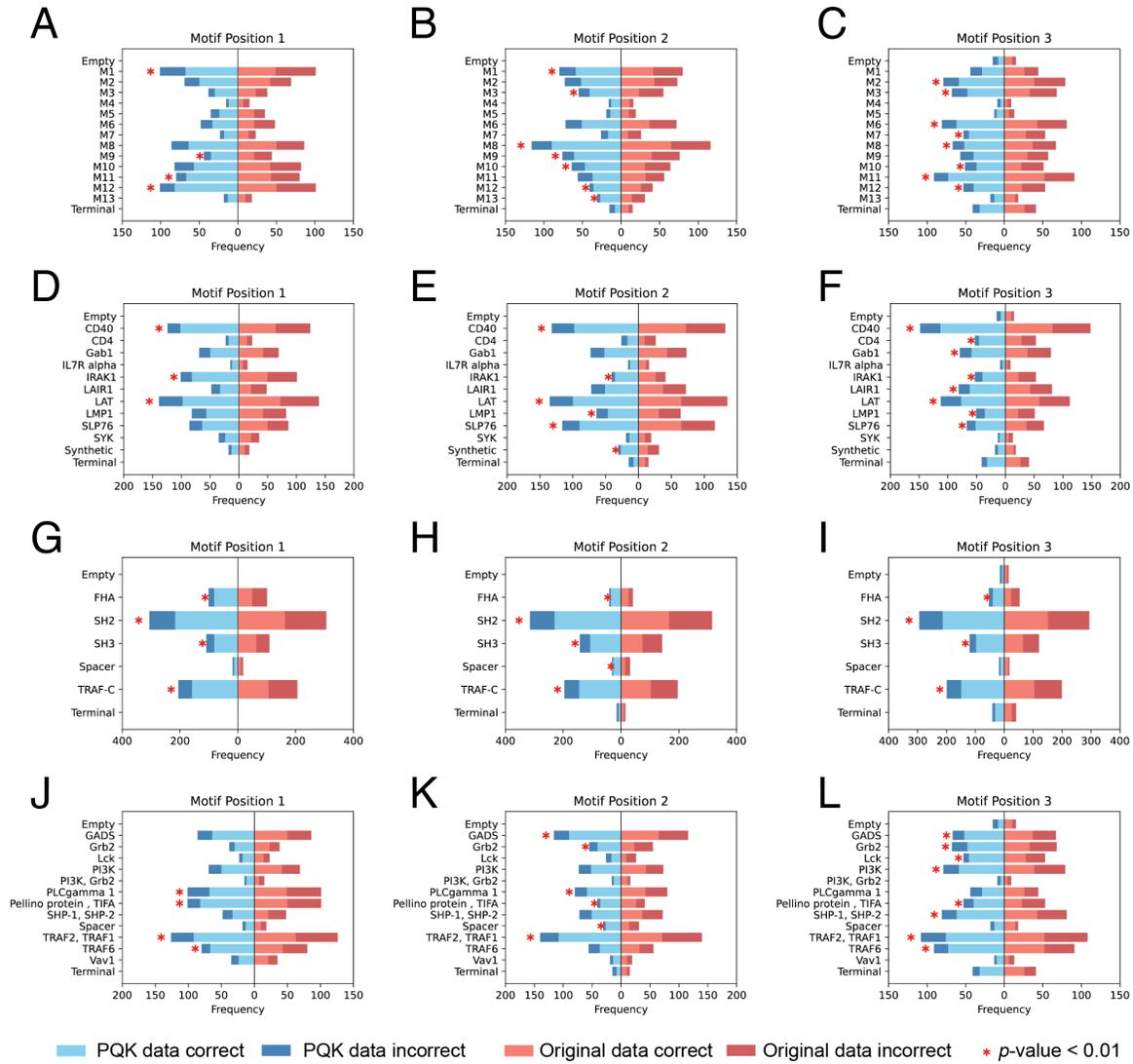

Figure 3. **Overall predictive performance per motif position**. The frequency of correctly and incorrectly predicted CAR T-cell test samples from across all 10 data splits are reported for PQK data (PQK) and original data (SVM) for each motif position and according to the specific signaling motif (A-C), source protein associated (D-F), signaling domain type (G-I), and binding partner (J-L). 'Empty' indicates padding entry needed for samples where there were no associated motifs in a given position. 'Terminal' indicates the terminal motif position. Fisher's exact test is used to determine whether a method was significantly better at predicting the given phenotype and a red asterisk indicates p-value < 0.01.

## Discussion

Our study represents one of the first large scale applications of QML to optimize cell therapeutic designs. Here we demonstrate that by transforming the dataset using PQK executed on a QPU, a classical SVM is able to achieve marginally improved performance with PQK transformed data compared to using the original dataset for classifying CAR T-cell cytotoxicity. Furthermore, the SVM with PQK transformed data demonstrates significantly improved predictive performance for particular motif types or in situations where information is constrained. Our prior work[12] has demonstrated that the current state-of-the-art classical ML methods are able to achieve only moderate performance ($R^2$=0.71) when regressing cytotoxicity. Here, the prior experimental design is reformulated as a classification task, and we show that in the PQK case we achieve a median F1 of 0.75 and maximal F1 of 0.81 across ten data splits as compared to a median F1 of 0.73 and maximal F1 of 0.77 achieved by the SVM on the original data alone.

While these F1 scores of quantum-enhanced data are only a slight improvement over the original data, we found that by using the PQK approach, the classifier is able to predict particular motifs and associated features at specific motif positions consistently, significantly better than the case where the original data is used. For instance, for motifs from source proteins CD40, IRAK1, and LAT, the predictive accuracy across all three motif positions of the classifier trained with PQK-data outperforms a classifier trained on the original data. This suggests that the PQK approach allows for a more accurate model for clinical use cases where motifs derived from those source proteins are preferred. This would be relevant, for example, to a clinician whose goal is to design the optimal CAR T-cell for antigen-low acute lymphoblastic leukemia patients where LAT derived motifs are preferred[41]. Therefore, PQK-based models would be preferred to purely classical kernel methods. Similarly, we find that in scenarios with lower amounts of information available, such as predicting CAR T-cells according to the motif in the third position where there are higher rates of 'empty' padding values or terminal motifs, PQK-based models are able to outperform classical SVM as demonstrated by the high rates of significantly improved predictive performance per motif, source protein, or binding partner. Here we show that for certain motifs, source proteins, or binding partners, as well as for cases where there is limited information, a quantum-classical kernel approach is able to achieve significantly improved performance which can potentially translate to meaningful clinical relevance. Our observations are also numerically supported by calculations of the geometric separation between the classical and projected quantum kernels and the corresponding model complexity metrics. Our computation of these metrics using initial small-scale circuits indicates that our dataset is in the regime where we can expect potential quantum advantage for prediction accuracy, which is further confirmed by our large-scale hardware experiments.

An intrinsic challenge for ML methods applied to problems such as CAR T-cell intracellular signaling domain optimization is the underdetermined nature of the problem. There is a lack of representative samples for the combinatorial space of available motifs, an issue not uncommon in many biological problems where specimens are often much fewer than the available feature space. There is a polynomial increase in motif combinations as the size of signaling domain libraries increases, and an exponential increase in combinations as the number of positions available in an engineered cell increases. These sample constraints make training generalizable, classical ML methods, whether SVMs or CNNs, a challenge. In this regard, PQK-based models may have some advantages as they have been shown to operate well for certain datasets[37], and QML in general has been shown to generalize better with few data[31].

Our study establishes the feasibility of PQK-based classification to aid in CAR T-cell design through improved predictive power, but we acknowledge the bulk of experiments focused on a reformulation of the original ML problem into classification task. We considered to instead perform a regression task, which the PQK workflow is able to support. However, given the current noise levels of near-term quantum hardware, regressing a robust, continuous value is a challenge. It would most likely require many replicated experiments to characterize the variability of the regressed value over multiple quantum projections of the same data. In the near term, until quantum error suppression and mitigation tools are further improved or as fault tolerant quantum devices come online as early as 2029[17], quantum ML algorithms may be better suited to classification tasks. Additional future work would be to apply a trained PQK model to predict the remainder of the combinatorial space of CAR T-cells from the given signaling motif library. This would enable the discovery of novel, effective CAR T-cells for experimental validation.

We believe that quantum algorithms, and hybrid quantum-classical machine learning approaches in particular, hold the potential for having a materially positive impact on CAR T-cell design. This would be of profound importance as the evidence for their efficacy grows beyond oncology to different application spaces from autoimmunity[42] to neurodegeneration[43]. The opportunity for PQK models such as those demonstrated here lies in identifying the problem areas where there are known bottlenecks in terms of predictive performance or classical computational bounds, and exploring the potential for this fundamentally different computing paradigm to make useful, significant biological insights.

## Method

### Datasets and preprocessing

The dataset used for this study is formed by 13 motifs (M1-M13), that could be combined in up to three different positions with all combinations ending with a final M14 terminal motif[12]. The authors measured experimentally the cytotoxicity and the stemness of the CAR T-cell populations of 246 constructs. Based on these experimental data points, they built a ML model to explore the entire combinatorial library and predict the phenotypes of the constructs not tested experimentally. We represented the distributions of the cytotoxicity and stemness experimental data in Supplementary Figure 6. The cytotoxicity data show a bimodal distribution (Supp. Fig. 6A), with a local minimum at 0.62. This profile enabled us to define two classes for CAR T-cell phenotypes, with CAR T-cell constructs showing high cytotoxicity behavior when Nalm6 survival is smaller than 0.62 and low cytotoxicity when Nalm6 survival is greater than 0.62. The distribution of the stemness experimental data exhibits one single asymmetric peak located at ~15 IL7RaKLRG1 (Supp. Fig. 6B), making the binary data classification infeasible for stemness. Therefore, we focused our study on a classification task predicting high and low CAR T-cell cytotoxicity.

## Data encoding strategies

We consider a total of 13 signaling motifs, M1 to M13, plus a terminating motif M14 over four positions and use a one-hot encoding strategy where the binary feature vector length equals to the number of observed motif classes in the experiment by the number of motif positions considered. We observe 15 classes for the 14 motifs (M1-14): 14 classes correspond to the motifs themselves, while the additional one accounts for an empty position, representing the absence of a motif in that position. This empty class was necessary in instances when sample data included only the first 1 or 2 motif positions with elements among the motifs M1-M13, followed by the terminating M14 motif, where 2 or 1 motif positions remain unoccupied, thus need to be filled with a placeholder class to ensure consistent input padding across all samples. Therefore, the largest experiments using all four positions would require at least 60 bits/qubits. Motif information[12] is included in Table 2.

*Table 2. Signaling motifs*

| Motif | Sequence (Core Motif) | Source Protein | Binding Partners | Binding Domain | Consensus motifs |
|---|---|---|---|---|---|
| M1 | DYHNPGYLVVLPDSTP | LAT | PLC$\gamma$1 | SH2 | Yx(A/I/L/V)(A/F/I/L/V/W/Y/P) |
| M2 | EELDENYVPMNPNSPP | Gab1 | PI3K | SH2 | YxxM |
| M3 | EEGAPDYENLQELNHP | LAT | Grb2 | SH2 | YXNX |
| M4 | LGSNQEEAYVTMSSFYQNQ | IL7R$\alpha$ | PI3K, Grb2 | SH2 | YxxM, YxNx |
| M5 | LPMDTEVYESPFADPEEIR | SYK | Vav1 | SH2 | Y(M/L/E)xP |
| M6 | KPMAESITYAAVARHSAG | LAIR1 | SHP-1, SHP-2 | SH2 | (S/I/V/L)xYxx(I/V/L) |
| M7 | LPTWSTPVQPMALIVLG | CD4 | Lck | SH3 | PxxPx(R/K) |
| M8 | PAPSIDRSTKPPLDRSL | SLP76 | GADS | SH3 | RxxK |
| M9 | GSNTAAPVQETLHGCQ | CD40 | TRAF2, TRAF1 | TRAF-C | Px(Q/E)E |
| M10 | DDSLPHPQQATDDSGHES | LMP1 | TRAF2, TRAF1 | TRAF-C | Px(Q/E)xxD, Px(Q/E)xT |
| M11 | KAPHAKQEPQEINFPDDLP | CD40 | TRAF6 | TRAF-C | PxExxZ |
| M12 | GSGPGSRPTAVEGLALGSS | IRAK1 | Pellino protein, TIFA | FHA | Txx(E/D), Txx(I/L/V) |
| M13 | SAGSAGSAGSAGSAGSAG | Synthetic | Non-functional spacer | | |

## Project quantum kernel method

In the Projected Quantum Kernel (PQK) method, data is first loaded onto a quantum circuit, otherwise referred to as quantum embedding. This quantum-embedded data now lies in a high-dimensional Hilbert space, which we project back into classical space by making measurements. These measurements are performed in all possible single-qubit Pauli bases to construct all single-qubit reduced density matrices (1-RDMs). 1-RDM features constitute the new dataset features on which classical kernel methods are applied.

When constructing quantum circuits for quantum embeddings, we select two embedding circuits proposed in Huang et al. [12,37]. The first embedding (E1) that we use is referred to as the ZZ Feature Map[38] in quantum information literature. Given that our dataset features consist of binary values, we use a scaling of either π or π/2 as the gate rotation angles.

$$U = exp\left(\sum_{j=0}^{n-1} x_{ij} Z_j + \sum_{j=0}^{n-1} \sum_{j'=0}^{n-1} Z_j Z_{j'}\right)$$

The second quantum embedding (E2) that we implement is a Hamiltonian evolution ansatz. In particular, we use the 1D-Heisenberg model Hamiltonian with 4 or 6 Trotter steps as our quantum embedding.

For the quantum projection experiments, projective measurements in the X, Y and Z basis, which determine the corresponding expectation values, were made. The 1-RDM for each qubit can then be determined by combining these expectation values. Therefore, we carry out three different experiments (one for each measurement basis) for each data point. In each experiment, we used 10000 shots, i.e. repeated executions of the quantum circuit, to calculate a statically significant estimate of the expectation value.

The experiments were performed on *ibm_marrakesh*, a superconducting qubit device with fixed frequency transmon qubits as data qubits connected via tunable couplers. The device is part of the family of the second revision of IBM Heron QPUs (R2), which features low-error two-qubit gate operations (0.3% median CZ error) with duration similar to that of single-qubit operations (0.02% median error) and low readout error (1% median error). To improve the quality of the obtained results we have utilized a combination of error suppression and mitigation techniques. A qubit layout selection routine was used to determine the sequence of best performing qubits for each experimental run. The qubits chosen in this way were then used in all experiments belonging to the same run. Pauli twirling[44] was used to tailor the noise in gate and measurement operations to Pauli noise. The 10000 shots taken for each measured expectation value were evenly distributed among the 64 twirled instances. Finally, TREX[45] was used to mitigate the readout error in the calculation of expectation values. The full experiment took 100 minutes of QPU time with each classical data point requiring ~22 seconds of quantum processing.

## Evaluating machine learning algorithms for cytotoxicity prediction

To evaluate the performance of the model, we generated 10 random 70/30 splits of the data into train and test sets, respectively, and tested two quantum embedding strategies. Due to the class imbalance in the dataset, we used the weighted F1 score as a suitable evaluation metric. To select which classical ML model works better, we evaluated the F1 of 27 different ML models without any parameter tuning using *lazypredict* (v0.2.16). SVM was selected for further hyperparameter optimization as it was one of the top performing methods from this analysis (see Supplementary Table S1); it was the reference classical kernel method throughout this study. *GridSearchCV* (in *scikit-learn* v1.6.0) was used to identify the optimal hyperparameters (see Table 3) to increase the model performance with 10 cross-validations of the training data. To ensure a proper comparison with and without the quantum embedded data, the same splits were used during the cross-validation step.

*Table 3. Hyperparameter tuning for SVM*

| Parameter | Values tested |
|---|---|
| kernel | {linear, poly, radial basis function, sigmoid} |
| C | {0.001, 0.005, 0.007, 0.01}<br>[0.01, 0.1] in increments of 0.01<br>[0.25, 14.75] in increments of 0.25<br>{20, 50, 100, 200, 500, 700, 1000, 1100, 1200, 1300, 1400, 1500, 1700, 2000} |
| gamma | {auto, scale, 0.001, 0.005, 0.007}<br>[0.01, 0.1] in increments of 0.01<br>[0.25, 14.75] in increments of 0.25<br>{20, 50, 100} |

## Evaluating effect of feature order on entanglement

To evaluate the potential impact of feature order in the quantum circuit on entanglement, we assess whether pre-processing the input features to arrange them according to their correlation would provide additional benefit. In the training data, we re-arrange data features by computing the Matthews correlation coefficient between all features and arrange them based on the dendrogram computed by hierarchical clustering with complete linkage. This arrangement of features is then provided to the quantum circuit for both the train and test data.

## Data Availability

In this study, we used publicly available datasets as detailed in Daniels et al.[12]

## Acknowledgements

The author would like to thank Kunal Sharma for his engaging dialogue on the original PQK paper, which significantly enriched the development of this research, and Prof. Kyle G. Daniels for insightful discussions on the data. Part of this work was supported by the National Science Foundation under Grant No. DBI-1548297.

## Author Contributions

Study conceptualization: F.U, K.R., T.G., S.C, L.P.; F.U., M.T. and M.A. performed all experiments. F.U. ran the machine learning pipeline advised by K.R.; M.T. implemented the PQK embedding and, with the assistance of T.G., performed the theoretical analysis; J.S. performed data preprocessing. M.A. assisted in conducting quantum hardware experiments and implemented the OBP; F.U. and K.R. interpreted the results and wrote the manuscript with M.T., S.C. and M.A., which was edited by the co-authors; L.P. supervised the work with the help of N.E.N.

## Ethics declarations

The other authors declare no competing interests.